\pdfoutput=1 
\documentclass[letterpaper, 10 pt, conference]{ieeeconf}
\usepackage{amsmath,amssymb,amsfonts}
\usepackage{graphicx}
\usepackage{booktabs}
\usepackage{hyperref}
\usepackage{xcolor}
\usepackage{microtype}
\usepackage{cite}

\title{\LARGE \bf Fiatlux: A Long-Horizon Benchmark for Humanoid Ladder Climbing and Light-Bulb Replacement}

\newcommand{\okina}{\raisebox{0.4ex}{\scalebox{0.7}{`}}}

\author{\authorblockN{Pavel Bushuyeu$^{*1}$, Yujin Chen$^{*1}$, Anton Nikolaev$^{*2}$, Brian Shu$^{*1,3}$, Igor Molybog$^{1}$}
\authorblockA{$^{1}$HawAII, University of Hawai\okina i at M\=anoa \quad $^{2}$Independent Researcher \quad $^{3}$Purdue University\\
{\small $^{*}$The first four authors contributed equally; their order is alphabetical by surname.}}}

\begin{document}

\maketitle

\begin{abstract}
Existing benchmarks evaluate tabletop manipulation, flat-floor household activity, or humanoid locomotion and manipulation as separate task groups; none scores vertical mobility and dexterous work on a fragile payload in one long-horizon episode. We present Fiatlux, a light-bulb replacement benchmark built on NVIDIA Isaac Lab. In one episode, a Unitree G1 humanoid positions a step ladder under a ceiling or wall fixture, climbs it, exchanges a spent bulb in a socket for a fresh one, and leaves the spent one in a disposal crate. We decompose the episode into twelve subtask environments scored on difficulty-weighted gates. The goal is a successful replacement, with the fresh bulb seated, the spent one disposed of, neither dropped, and a fragility bound not crossed. Runs that fall short can earn partial credit. Observations are split into a standard mode (signals a physical robot could sense or estimate) and a privileged mode (exact simulator state). We specify the evaluation protocol and provide reference baseline implementations spanning RSL-RL PPO, zero-shot NVIDIA GR00T N1.7 Vision-Language-Action (VLA) models, and whole-body controllers. Additionally, we provide the teleoperated recordings used to specify and check the success gates. The benchmark code and the teleoperated recordings are available at \href{https://fiatlux-bench.github.io}{fiatlux-bench.github.io}.
\end{abstract}


\section{Introduction}

GPU-accelerated simulation platforms enable whole-body locomotion and complex object manipulation \cite{makoviychuk2021isaac, mittal2023orbit}. However, existing benchmarks keep the robot on the ground, from tabletop and mobile-base manipulation \cite{yu2020meta, gu2023maniskill2} to everyday activities in furnished scenes \cite{srivastava2022behavior, nasiriany2024robocasa}. Bipedal humanoids can remove that constraint, so that a single task can span locomotion, ladder climbing, dual-arm manipulation, and precise object seating under safety constraints. Deploying humanoid robots for those tasks could reduce risk by isolating operators from electrical hazards and high falls. However, even a task such as ascending an A-frame step ladder requires three-dimensional center-of-mass (CoM) balance, multi-limb contact transitions, rung force management, and tilt stabilization, which makes it a demanding target for whole-body reinforcement learning and Vision-Language-Action (VLA) models.

To measure progress on those challenges, we introduce Fiatlux, an open-source, reproducible benchmark built as a native extension for NVIDIA Isaac Lab. Fiatlux evaluates humanoid ladder climbing and light-bulb replacement using a Unitree G1 humanoid platform, with task assets drawn from the NVIDIA Omniverse libraries. The core contributions of Fiatlux encompass:

\begin{itemize}
    \item \textbf{Benchmark for Climb-Required Maintenance}: Formalizes overhead bulb replacement as a task family in which the fixture sits above the robot's standing reach, so the episode cannot be completed without climbing.
    \item \textbf{Modular Subtask Architecture and Unified Scene Hierarchy}: Decomposes multi-stage execution into 12 structured atomic subtasks ($S_{01}$--$S_{12}$) categorized across Flat Navigation, Manipulation, and Climbing control modes, sharing a unified family scene hierarchy with per-task preset layouts. Each subtask is a registered environment with a start pose approximating the state its predecessor ends in, and its own success gate defining a progress boundary.
    \item \textbf{Automated Asset Generation}: Converts object images into textured USD assets through a staged image-to-mesh pipeline with optional articulation prediction and a structural verifier, supporting benchmark extension through explicit calibration rather than hand-modeled meshes.
    \item \textbf{Domain Randomization}: Randomizes the scene layout per layout draw, grip friction at startup, and lighting, material colors, and the robot start state per reset.
    \item \textbf{Teleoperated Demonstration Adapters}: Provides a teleoperation adapter for every subtask, in which an operator in a virtual-reality headset drives the arms through inverse kinematics while the GEAR-SONIC controller walks and balances the legs.
    \item \textbf{Empirical Baseline Evaluation}: Establishes zero-shot evaluation protocols for Vision-Language-Action (VLA) architectures, tested specifically on NVIDIA GR00T N1.7 \cite{gr00t2024nvidia} paired with whole-body controllers.
\end{itemize}

The rest of this paper is organized as follows. Section~\ref{sec:background} reviews background literature in humanoid simulation, ladder climbing, the robot platform, and baseline controllers. Section~\ref{sec:method} details the task taxonomy, observation and action spaces, the climbing MDP, reward mechanics, success and safety metrics, domain randomization and teleoperation. Section~\ref{sec:results} gives the evaluation protocol, the released baselines, and how task achievability is established. Section~\ref{sec:discussion} discusses sim-to-real implications and the benchmark's current limitations.

\section{Background and Related Work}
\label{sec:background}

\subsection{Benchmark Frameworks}

Standard manipulation benchmarks like Meta-World \cite{yu2020meta} and ManiSkill \cite{gu2023maniskill2} focus on tabletop robotic arms. BEHAVIOR-1K \cite{srivastava2022behavior} introduces household activities; RoboCasa \cite{nasiriany2024robocasa} and RoboCasa365 \cite{nasiriany2026robocasa365} scale kitchen manipulation, the former across over 150 object categories and 100 evaluation tasks. HumanoidBench \cite{sferrazza2024humanoidbench} presents a simulated humanoid benchmark combining 15 whole-body manipulation and 12 locomotion tasks across a 61-DoF action space, yet evaluates each task in isolation. These frameworks leave an exposed gap in climbing during manipulation and scoring of a fragile payload. Fiatlux closes it in Isaac Lab, in one episode where the robot places its own ladder, climbs it, and exchanges a bulb under a force bound at the hands.

\subsection{Whole-Body Humanoid Ladder Climbing}
Ascending an A-frame step ladder couples lower-limb stepping with upper-limb contact stabilization. Unlike flat-ground walking, it constrains where the feet can land and demands a sustained forward lean.

Humanoid-Gym \cite{gu2024humanoidgym} establishes an Isaac Gym framework for bipedal locomotion with zero-shot sim-to-real transfer, but it addresses locomotion only. Model-free algorithms struggle at this action-space scale \cite{sferrazza2024humanoidbench}, which motivates hierarchical control. Puppeteer \cite{hansen2024puppeteer} pairs a latent-space world model (TD-MPC2 \cite{hansen2024tdmpc2}) with a low-level tracking policy trained on motion capture, which yields more natural whole-body motion. Closest to this work is LadderMan \cite{zhao2026ladderman}, which climbs an inclined ladder on a Unitree G1 from a depth-based visuomotor policy and demonstrates on-ladder manipulation, including bulb handling, under teleoperation. LadderMan is a control method validated by hardware demonstration; Fiatlux is a benchmark that scores a similar workflow as autonomous subtasks. Where LadderMan teleoperates the on-ladder work on a pre-placed ladder, Fiatlux scores the whole errand, from positioning a free-standing ladder to disposing of the spent bulb, and bounds the force the hands may apply to the payload.

\subsection{Humanoid Hardware Platform and Actuator Modeling}
The target hardware platform in Fiatlux is the Unitree G1 humanoid robot. The robot has a 29-DoF articulated body with dual multi-fingered hands:

\begin{itemize}
    \item \textbf{Inspire Hand Variant}: Multi-finger hand with position limits and torque capped at $0.5\text{ N}\cdot\text{m}$ for hardware safety.
    \item \textbf{Dex3 Hand Variant}: Alternative dexterous hand with seven joints per hand and its own finger torque cap.
\end{itemize}

Actuator modeling in Fiatlux reflects physical motor characteristics. Joint drives use proportional-derivative (PD) position controllers with stiffness ($K_p$) and damping ($K_d$) gains matched to G1 specifications. Armature inertias are set per Unitree motor type.

\subsection{Vision-Language-Action Models and Controllers}
Vision-Language-Action (VLA) models are the current candidate for policies that take a task as language and act from vision, which is the setting this benchmark targets. NVIDIA GR00T N1.7 \cite{gr00t2024nvidia} is a VLA that pairs a vision-language module with a diffusion transformer action head. Action Chunking with Transformers (ACT) \cite{zhao2023act} predicts a whole action sequence at once, Diffusion Policy \cite{chi2023diffusionpolicy} denoises a sequence over many steps, and $\pi_0$ \cite{black2024pi0} takes ten flow-matching steps, reaching the 50 Hz rate this benchmark also runs at. Additionally, PointAction \cite{wang2026pointaction} predicts future RGB frames and 3D point dynamics as an embodiment-agnostic action interface, which a decoder maps to robot actions. Fiatlux includes GR00T N1.7 as a zero-shot baseline (Section~\ref{sec:results}).

A VLA like GR00T N1.7 does not drive a humanoid's legs, so the GEAR Whole-Body Controller (WBC) turns the VLA's navigation, height, and torso commands into 50 Hz leg and waist control, while its arm targets apply directly.

\section{Methodology and Implementation}
\label{sec:method}

\subsection{Task Taxonomy and Control Mode Decomposition}
Fiatlux structures the long-horizon light-bulb replacement task into a single scene built in NVIDIA Isaac Lab on the PhysX solver. The complete benchmark task evaluates the multi-stage pipeline as one flat episode ($T = 72{,}000$ steps at 50\,Hz, a $1{,}440\text{ s}$ horizon). To establish clear, non-overlapping operational boundaries for stage-wise policy evaluation, Fiatlux registers 12 atomic subtasks as separate environments (Table~\ref{tab:fiatlux_subtasks}).

Every subtask is named by its dominant control mode:
\begin{enumerate}
    \item \textbf{Flat Navigation}: Whole-body navigation and locomotion across horizontal floor surfaces, including walking with and without payloads (ladder or bulb), and positioning the step ladder ($S_{01}, S_{05}, S_{07}, S_{09}$).
    \item \textbf{Manipulation}: Fine arm-and-hand trajectory execution, including bulb extraction from the fixture, bulb release into the disposal crate, fresh bulb grasping, and seating the fresh bulb ($S_{03}, S_{06}, S_{08}, S_{11}$).
    \item \textbf{Climbing}: Multi-limb contact transitions, vertical center-of-mass (CoM) balance management, and ascending or descending the ladder steps with and without payloads ($S_{02}, S_{04}, S_{10}, S_{12}$).
\end{enumerate}

The modes are not exclusive. $S_{03}$ and $S_{11}$ extend the on-ladder balance with manipulation. $S_{04}$ and $S_{10}$ combine climbing with constraints related to handling a glass bulb. $S_{02}$ and $S_{12}$ are the only climbing subtasks with the hands free. A policy competent at each mode separately can still fail where both run at once.

\begin{table*}[t]
\centering
\caption{Taxonomy of the 12 atomic subtasks ($S_{01}$--$S_{12}$) in Fiatlux across the three control modes. Four subtasks run two modes at once, dominant first. Each evaluation criterion is a conjunction whose clauses must hold at the same instant; rows marked sustained must hold continuously for the stated window. Throughout, robot standing means pelvis height above $0.35\text{ m}$ and tilt under $1.0\text{ rad}$, ladder vertical means tilt under $0.6\text{ rad}$, held and released mean grip contact force above and below $1.0\text{ N}$, and at rest means below $0.05\text{ m/s}$ and $0.10\text{ rad/s}$.}
\label{tab:fiatlux_subtasks}
\resizebox{\textwidth}{!}{%
\begin{tabular}{c l l p{10.5cm}}
\toprule
\textbf{ID} & \textbf{Subtask Name} & \textbf{Control Mode} & \textbf{Quantitative Evaluation / Termination Criterion} \\
\midrule
$S_{01}$ & Move Ladder & Flat Navigation & Ladder top within $0.67\text{ m}$ of the fixture, upright, ladder feet within $0.02\text{ m}$ of the floor, ladder at rest, robot standing (sustained $1.0\text{ s}$) \\
$S_{02}$ & Climb Ladder & Climbing & Pelvis no more than $0.15\text{ m}$ below the ladder's live top point and within $0.6\text{ m}$ of it horizontally, speed $< 1.5\text{ m/s}$, robot standing, ladder vertical \\
$S_{03}$ & Remove Old Bulb & Manipulation, Climbing & Old bulb released by the socket and $0.10\text{ m}$ clear of the seat, held, elevation $\ge 0.15\text{ m}$, robot standing, ladder vertical (sustained $0.5\text{ s}$) \\
$S_{04}$ & Descend with Bulb & Climbing, Manipulation & Pelvis below floor-stance height, within $0.6\text{ m}$ of the ladder, speed $< 1.5\text{ m/s}$, bulb held and elevated, robot standing, ladder vertical \\
$S_{05}$ & Carry Bulb to Disposal & Flat Navigation & Within $0.50\text{ m}$ of the crate's footprint, facing it within $0.5\text{ rad}$, speed $< 1.0\text{ m/s}$, bulb held \\
$S_{06}$ & Dispose Bulb & Manipulation & Old bulb geometry fully inside the crate interior and below its rim, translating below $0.05\text{ m/s}$, grip released, robot standing (sustained $1.0\text{ s}$) \\
$S_{07}$ & Approach New Bulb & Flat Navigation & Within $0.63\text{ m}$ of the fresh bulb, a bench-clearing standoff that leaves it outside the $0.50\text{ m}$ horizontal reach, facing it within $0.5\text{ rad}$, speed $< 1.0\text{ m/s}$ \\
$S_{08}$ & Grab New Bulb & Manipulation & Fresh bulb lifted $0.03\text{ m}$ off the bench, at least two bodies of one hand in contact above $1.0\text{ N}$, no hand body above $50\text{ N}$, robot standing (sustained $0.5\text{ s}$) \\
$S_{09}$ & Carry Bulb to Ladder & Flat Navigation & Within mounting range of the ladder, facing it within $0.5\text{ rad}$, speed $< 1.0\text{ m/s}$, bulb held, ladder vertical \\
$S_{10}$ & Climb with Bulb & Climbing, Manipulation & Top stance as $S_{02}$, bulb held and elevated, robot standing, ladder vertical \\
$S_{11}$ & Screw in Bulb & Manipulation, Climbing & Fresh bulb attached in the fixture, settled in position and orientation, grip released, robot standing, ladder vertical (sustained $1.0\text{ s}$) \\
$S_{12}$ & Climb Down & Climbing & Floor stance as $S_{04}$, fresh bulb still seated in the fixture, robot standing, ladder vertical \\
\bottomrule
\end{tabular}%
}
\end{table*}

All subtasks share one scene configuration, which sets the robot's joint drives, camera mounts, and contact sensors. Each subtask is a registered environment with a $120\text{ s}$ episode horizon, starting from a state designed to approximate where its predecessor ends, down to what the robot is already holding. 

\subsection{Observation Spaces}
\label{subsec:obs}
The benchmark separates what a real robot could measure from privileged ground-truth simulation state. Let $\mathcal{S}$ denote the state space of the simulator and $\mathbf{s}_t \in \mathcal{S}$ represent the full environment state at time step $t$. Fiatlux defines two observation mappings.

\subsubsection{Standard Observation}
The \textit{standard observation} mapping $h_{\text{std}}: \mathcal{S} \to \mathcal{O}_{\text{standard}}$ produces an observation vector $\mathbf{o}_t^{\text{std}} \in \mathcal{O}_{\text{standard}}$ composed of signals a physical robot could sense or estimate:
\begin{multline}
\mathbf{o}_t^{\text{std}} = \Big[ \mathbf{q}_t, \, \dot{\mathbf{q}}_t, \, \boldsymbol{\omega}_{\text{base}}, \, \mathbf{g}_{\text{proj}}, \, \hat{z}_{\text{base}}, \, \hat{\mathbf{v}}_{\text{base}}, \\ \mathbf{F}_{\text{contact}}, \, \boldsymbol{\psi}_{\text{rgb}}, \, \mathbf{d}_{\text{lidar}} \Big]^T
\end{multline}
where:
\begin{itemize}
    \item $\mathbf{q}_t \in \mathbb{R}^{d_{\text{joint}}}$ and $\dot{\mathbf{q}}_t \in \mathbb{R}^{d_{\text{joint}}}$ represent joint position and velocity vectors relative to default posture across $d_{\text{joint}} = 53$ actuated degrees of freedom, 29 body joints plus 12 in each of the two Inspire hands.
    \item $\boldsymbol{\omega}_{\text{base}} \in \mathbb{R}^3$ denotes base angular velocity measured by the robot Inertial Measurement Unit (IMU).
    \item $\mathbf{g}_{\text{proj}} \in \mathbb{R}^3$ is the gravity vector projected into the robot base coordinate frame $\mathcal{F}_{\text{base}}$.
    \item $\hat{z}_{\text{base}} \in \mathbb{R}$ and $\hat{\mathbf{v}}_{\text{base}} \in \mathbb{R}^3$ are base height and base linear velocity from the state estimator.
    \item $\mathbf{F}_{\text{contact}} \in \mathbb{R}^{3B}$ holds one world-frame force vector per body. The $B = 34$ bodies span both hands and wrists, as either hand can crush the bulb. The sensor registers only contact with the two bulbs, disregarding scenery the arm rests against and the robot's own body. The climbing task of Section~\ref{subsec:climb} substitutes contact at the feet and palms against the ladder. The palms are measured directly. The G1 has no foot sensor, so foot contact is estimated from ankle joint torques, in simulation as on hardware, and the policy consumes the same signal in both.
    \item $\boldsymbol{\psi}_{\text{rgb}} \in \mathbb{R}^{d_{\text{rgb}}}$ is the visual feature vector ($d_{\text{rgb}} = 512$) extracted via ResNet-18 from the head-mounted RGB camera.
    \item $\mathbf{d}_{\text{lidar}} \in \mathbb{R}^{d_{\text{lidar}}}$ contains range measurements across $d_{\text{lidar}} = 5,760$ raycast channels ($32$ elevation rows $\times$ $180$ azimuth samples) from the head-mounted Mid-360 LiDAR scanning ground and step-ladder geometry.
\end{itemize}
Sensor noise is declared per term and is additive uniform on a symmetric interval. A corrupted term $i$ is read as $\mathbf{o}_i + \boldsymbol{\epsilon}_i$ with $\boldsymbol{\epsilon}_i \sim \mathcal{U}(-\delta_i, \delta_i)$. The half-widths $\delta_i$ are $0.2$ for base angular velocity, $0.05$ for projected gravity, $0.1$ for estimated base linear velocity, $0.05$ for estimated base height, $0.01$ for joint position, and $1.5$ for joint velocity. The contact, camera, and LiDAR terms are read uncorrupted, so the exteroceptive channels of the standard mode carry no sensor model.

\subsubsection{Privileged Observation}
The \textit{privileged observation} mapping $h_{\text{priv}}: \mathcal{S} \to \mathcal{O}_{\text{privileged}}$ constructs an uncorrupted state vector $\mathbf{o}_t^{\text{priv}} \in \mathcal{O}_{\text{privileged}}$ for critic training and privileged baselines:
\begin{multline}
\mathbf{o}_t^{\text{priv}} = \Big[ \mathbf{T}_{\text{robot}}, \, \mathbf{T}_{\text{ladder}}, \, \mathbf{T}_{\text{fixture}}, \, \mathbf{T}_{\text{fresh}}, \\ \mathbf{T}_{\text{old}}, \, \mathbf{T}_{\text{crate}}, \, \mathbf{d}_{\text{score}}(t), \, \mathbf{b}_{\text{seat}}(t) \Big]^T
\end{multline}
where $\mathbf{T}_i = (\mathbf{p}_i, \mathbf{r}_i) \in \mathbb{SE}(3)$ denotes the absolute 6D pose of the robot, ladder, fixture, fresh bulb, old bulb, or disposal crate, with translation $\mathbf{p}_i \in \mathbb{R}^3$ and unit quaternion $\mathbf{r}_i \in \mathbb{SO}(3)$. The vector $\mathbf{d}_{\text{score}}(t) = [d_1(t), d_2(t), d_3(t), d_4(t)]^T \in \mathbb{R}^4$ captures the four score-relevant spatial distances:
\begin{align}
d_1(t) &= \|\mathbf{p}_{\text{ladder\_top}}(t) - \mathbf{p}_{\text{fixture\_seat}}(t)\|_2, \\
d_2(t) &= \|\mathbf{p}_{\text{fresh\_plug}}(t) - \mathbf{p}_{\text{fixture\_seat}}(t)\|_2, \\
d_3(t) &= \|\mathbf{p}_{\text{old\_plug}}(t) - \mathbf{p}_{\text{fixture\_seat}}(t)\|_2, \\
d_4(t) &= \|\mathbf{p}_{\text{old\_bulb}}(t) - \mathbf{p}_{\text{crate}}(t)\|_2.
\end{align}
The final entry $\mathbf{b}_{\text{seat}}(t) \in \{0, 1\}^2$ reports whether the old and the fresh bulb are currently seated in the fixture. It is privileged as no sensor can distinguish a bulb the socket is holding from one resting in the same position. The two groups are disjoint, and a critic that needs both reads their concatenation.

\subsection{Action Parameterization}
The action space $\mathcal{A} = [-1, 1]^{d_{\text{act}}}$ parameterizes joint-position targets at a 50\,Hz control rate, where $d_{\text{act}} = d_{\text{joint}} = 53$. At step $t$, the policy outputs an action vector $\mathbf{a}_t \in \mathcal{A}$. Target joint angles $\mathbf{q}_{\text{cmd}, t} \in \mathbb{R}^{d_{\text{joint}}}$ are computed relative to the standing posture vector $\mathbf{q}_0 \in \mathbb{R}^{d_{\text{joint}}}$:
\begin{equation}
\mathbf{q}_{\text{cmd}, t} = \mathbf{q}_0 + \mathbf{S}_{\text{act}} \mathbf{a}_t
\end{equation}
where $\mathbf{S}_{\text{act}} = \operatorname{diag}(s_1, \dots, s_{d_{\text{act}}}) \in \mathbb{R}^{d_{\text{act}} \times d_{\text{act}}}$ contains joint action scaling coefficients, uniform at $s_i = 0.5$ across the joint set. Since the offset is the standing posture, the zero action commands it. Low-level proportional-derivative (PD) motor drives compute joint torque commands $\boldsymbol{\tau}_t \in \mathbb{R}^{d_{\text{joint}}}$ at the $200\text{ Hz}$ physics rate, four steps per action:
\begin{equation}
\boldsymbol{\tau}_t = \mathbf{K}_p (\mathbf{q}_{\text{cmd}, t} - \mathbf{q}_t) - \mathbf{K}_d \dot{\mathbf{q}}_t
\end{equation}
where $\mathbf{K}_p, \mathbf{K}_d \in \mathbb{R}^{d_{\text{joint}} \times d_{\text{joint}}}$ are diagonal stiffness and damping matrices. These are implicit actuators, so PhysX evaluates the law inside the solver and the environment writes no torques. The applied torque is clipped to each joint's effort limit. The previous action $\mathbf{a}_{t-1}$ is fed back into the policy input of Section~\ref{subsec:obs}. The controller knows what it last commanded, and that command carries actuator state the joint sensors do not expose.

\subsection{Step-Ladder Climbing Task MDP}
\label{subsec:climb}
The standalone step-ladder climbing task evaluates ascent in an environment separate from the climbing subtasks of Table~\ref{tab:fiatlux_subtasks}. The robot starts facing an A-frame step ladder ($1.86\text{ m}$ tall, $0.61 \times 0.98\text{ m}$ footprint). Episodes run for $120\text{ s}$.

\subsubsection{Climbing Action and Observation Specifications}
The action space is the whole-body joint set ($d_{\text{act}} = 53$), including leg, waist, arm, and finger joint targets. The two palms report contact force directly, and each foot's contact is estimated from its ankle joint torques, the estimate a physical G1 would run. Both register only contact with the ladder, disregarding the floor and the robot's own body. Stacking the four world-frame forces gives $\mathbf{F}_{\text{rung}} = [\mathbf{f}_{\text{lf}}, \mathbf{f}_{\text{rf}}, \mathbf{f}_{\text{lp}}, \mathbf{f}_{\text{rp}}]^T \in \mathbb{R}^{12}$, where each $\mathbf{f}_i \in \mathbb{R}^3$ is the force on one limb, the left and right foot and the left and right palm.

\subsubsection{Climbing Reward Mechanics}
The climbing reward pays height gain and rung contact, awards a bonus on success, and penalizes instability and rough motion.
\begin{multline}
R_{\text{climb}}(t) = w_z \Delta r_{\text{climb}}(t) + w_c f_{\text{contact}}(t) + C_{\text{climb}} \mathbb{I}_{\text{climbed}}(t) \\ - \lambda_{\text{sway}} P_{\text{sway}}(t) - \lambda_{\text{wobble}} P_{\text{wobble}}(t) + R_{\text{shape}}(t)
\end{multline}
where:
\begin{itemize}
    \item Progressive height gain $\Delta r_{\text{climb}}(t)$ pays new maximum root heights reached during the episode:
    \begin{multline}
    \Delta r_{\text{climb}}(t) = \max \Big( 0, \, \max_{0 \le \tau \le t} z_{\text{root}}(\tau) \\ - \max_{0 \le \tau \le t-1} z_{\text{root}}(\tau) \Big)
    \end{multline}
    \item Limb contact fraction $f_{\text{contact}}(t) = \frac{1}{4} \sum_{i=1}^4 \mathbb{I}(\|\mathbf{F}_{\text{rung}, i}(t)\|_2 > F_{\text{min}})$ encourages rung contact.
    \item CoM sway penalty $P_{\text{sway}}(t) = \|\mathbf{v}_{\text{CoM}, xy}(t)\|_2^2$ penalizes horizontal velocity drift.
    \item Angular rate penalty $P_{\text{wobble}}(t) = \|\boldsymbol{\omega}_{\text{base}, xy}(t)\|_2^2$ penalizes roll and pitch wobble.
\end{itemize}
Height gain and the success bonus both weigh $500$ and dominate the reward. Rung contact weighs $0.25$ and counts a limb as in contact above $F_{\text{min}} = 1.0\text{ N}$. Stability shaping is smaller still, $-0.5$ on sway and $-0.05$ on wobble. Of the six motion-shaping penalties only one is large, the $-200$ charged when a fall ends the episode. The rest run from $-1.0$ for ankle position limits down to $-1.25 \times 10^{-7}$ for hip and knee acceleration, with action rate, waist deviation, and finger deviation in between.

\subsubsection{Climbing Success and Fall Predicates}
\begin{itemize}
    \item \textbf{Climbing Success Predicate} $\mathbb{I}_{\text{climbed}}(t)$: Triggers when $z_{\text{root}} > 1.82\text{ m}$, $\|\mathbf{p}_{\text{root}, xy} - \mathbf{p}_{\text{ladder}, xy}\|_2 < 0.60\text{ m}$, and $\|\mathbf{v}_{\text{root}}\|_2 < 1.50\text{ m/s}$, so the robot must be at working height on the ladder and settled, not passing through.
    \item \textbf{Climbing Fall Termination} $\mathbb{I}_{\text{fall}}(t)$: Triggers when $z_{\text{root}} < 0.35\text{ m}$ or $\theta_{\text{tilt}} > 1.0\text{ rad}$.
\end{itemize}

\subsection{Reward Mechanics and Normalized Progress}
The benchmark task's per-step progress rewards use potential-based shaping \cite{ng1999policy}. Let $d^{(g)}(t)$ denote the distance to the target state for subgoal $g$, mirroring the four score-relevant distances of Section~\ref{subsec:obs}, with $d_0^{(g)} = d^{(g)}(0)$ captured after reset randomization and floored at $10^{-3}\text{ m}$. The potential $\Phi_g$ is the normalized progress ratio, with $\operatorname{clamp}(x, a, b) = \min(\max(x, a), b)$:
\begin{equation}
\Phi_g(t) = \operatorname{clamp}\left( \frac{d_0^{(g)} - d^{(g)}(t)}{d_0^{(g)}}, \, 0, \, 1 \right)
\end{equation}
Normalizing by the episode's own start distance makes arrival worth $1$ from any spawn, so a lucky starting layout cannot outscore an unlucky one.

Each step pays the change in potential,
\begin{equation}
\Delta r_{\text{prog}, g}(t) = \Phi_g(t) - \Phi_g(t-1),
\end{equation}
so approaching pays positive, retreating pays negative, and holding position pays nothing. Over an episode the per-step payments cancel in pairs, leaving $\Phi_g(T) - \Phi_g(0)$, which reports where the episode ended, not the best it ever reached, while each step still carries a dense gradient. The climbing task of Section~\ref{subsec:climb} pays its height term by a different rule, tracking the episode's best height and paying each improvement to the record. Under that rule a policy could carry the bulb almost to the crate, drop it there, and keep nearly full credit for a disposal that never happens. The height term, by contrast, pays each new centimeter exactly once, so re-climbing old ground earns nothing, and a fall is charged by its own penalty and ends the episode, not by refunding the climb.

Old-bulb removal measures progress away from the seat, not toward it. The bulb starts seated, so $d_0$ is near zero. Its potential is instead the clearance achieved, measured against an absolute threshold:
\begin{equation}
\Phi_{\text{old}}(t) = \operatorname{clamp}\left( d^{(\text{old})}(t) \, / \, d_{\text{clear}}, \, 0, \, 1 \right), \quad d_{\text{clear}} = 0.10\text{ m}
\end{equation}
The episodic reward combines the four progress terms with completion bonuses $C_j$ and penalties $P_k$:
\begin{align}
R(t) = {} & w_{\text{lad}} \Delta r_{\text{prog}, \text{ladder}}(t) + w_{\text{fresh}} \Delta r_{\text{prog}, \text{fresh}}(t) \nonumber \\
& + w_{\text{old}} \Delta r_{\text{prog}, \text{old}}(t) + w_{\text{disp}} \Delta r_{\text{prog}, \text{disposal}}(t) \nonumber \\
& + \sum_{j} C_j \mathbb{I}_j(t) - \sum_{k} \lambda_k P_k(t)
\end{align}
Each term pays value $\times$ weight $\times$ $dt$ per step, with $dt = 0.02\text{ s}$. A progress potential saturates at $1$ and a completion bonus pays once, so a weight of $500$ contributes at most $10$ over an episode, independently of the horizon:
\begin{itemize}
    \item The four progress terms are bounded at $5$ for the ladder, $10$ for the fresh bulb, $5$ for old-bulb removal, and $10$ for disposal.
    \item The completion bonuses $C_j$ pay $2$ for the ladder in place, $5$ for the fresh bulb inserted, $3$ for the old bulb removed, $5$ for it disposed, and $10$ for full success.
    \item The penalties $P_k$ charge falling, tipping the ladder, dropping either bulb, and a set of stability and smoothness terms, each sized well below the bounds above.
\end{itemize}
A perfect run therefore scores at most $55$. A completion bonus pays on the first step its predicate holds. A seated bulb stays seated, so paying the raw predicate would reward occupying an achieved state until the horizon expires instead of finishing the task and ending the episode. Every reward term is logged separately, so the per-term episode sums are the score breakdown.

\subsection{Success, Termination, and Safety Metrics}
Seat admission, task success, and early termination are per-step predicates:

\begin{itemize}
    \item \textbf{Seat Admission Predicate} $\mathbb{I}_{\text{seat}}(t)$: a free bulb becomes seated when its plug reaches the socket seat within $0.004\text{ m}$ along the seat axis and $0.015\text{ m}$ across it, with the plug-to-seat axis tilt under $0.2\text{ rad}$ and the socket unoccupied by the other bulb. Rotation about the mating axis is unconstrained. A seated bulb is then held like a magnetic snap fitting, by a continuously applied retention wrench (mimicking magnetic lock), not by the bore's geometry, and is freed only by a sustained axial withdrawal past $0.015\text{ m}$, so a momentary knock does not unseat it.
    \item \textbf{Task Success Predicate} $\mathbb{I}_{\text{success}}(t) = \mathbb{I}_{\text{seat}}^{\text{fresh}}(t) \land \mathbb{I}_{\text{disposed}}^{\text{old}}(t)$: the fresh bulb is seated in the fixture \emph{and} the released old bulb rests inside the disposal crate, by the same crate-interior test as the $S_{06}$ row of Table~\ref{tab:fiatlux_subtasks}. Seating alone does not end the episode successfully. A run that installs the fresh bulb correctly and leaves the spent one on the floor scores zero, making disposal part of the task, not an epilogue to it.
    \item \textbf{Robot Fall Termination} $\mathbb{I}_{\text{fall}}(t)$: Triggers at $z_{\text{root}} < 0.35\text{ m}$, well below the $0.79\text{ m}$ standing pelvis height, or at torso tilt $\theta_{\text{tilt}} > 1.0\text{ rad}$.
    \item \textbf{Ladder Tip Termination} $\mathbb{I}_{\text{tip}}(t)$: Triggers at $\theta_{\text{ladder}} > 0.6\text{ rad}$, the ladder's tilt from vertical.
    \item \textbf{Bulb Drop Termination} $\mathbb{I}_{\text{drop}}(t)$: Triggers at $z_{\text{fresh}} < 0.40\text{ m}$ for the fresh bulb, or at $z_{\text{old}} < 0.15\text{ m}$ for the released old bulb while outside the crate. A correctly disposed bulb also comes to rest near the floor, so without the crate exclusion disposal would score as a drop.
\end{itemize}

\subsubsection{Offline Bag Scoring Protocol}
Scoring is decoupled from the simulator. It reads a recorded trajectory bag and derives every number from recorded state, so one bag can be rescored under different thresholds without running the simulation again. An episode is classified as \textit{broken} if the peak contact force over both hands, over all steps and all hand bodies, exceeds the fragility limit:
\begin{multline}
\mathbb{I}_{\text{broken}} = \mathbb{I} \left( \max_{0 \le t \le T} \|\mathbf{F}_{\text{hand}}(t)\|_2 > F_{\text{fragile}} \right), \\ F_{\text{fragile}} = 50.0\text{ N}
\end{multline}
Both hands are read since the operator uses whichever is convenient, and scoring one arm records a left-handed crush as clean. Finger torque saturates at $0.5\text{ N}\!\cdot\!\text{m}$, which stalls a closing finger at the bulb surface at roughly $12\text{ N}$ of grip, so the $35\text{ g}$ bulb can be held and released well inside the bound. The headline metric is the Clean Success Rate ($\text{CSR}$) over $N$ evaluation episodes:
\begin{equation}
\text{CSR} = \frac{1}{N} \sum_{i=1}^{N} \left( \mathbb{I}_{\text{success}}^{(i)} \land \neg \mathbb{I}_{\text{broken}}^{(i)} \land \neg \mathbb{I}_{\text{drop}}^{(i)} \right)
\end{equation}
Since $\mathbb{I}_{\text{success}}$ is the full-replacement predicate, a clean success is an episode that seated the fresh bulb, disposed of the spent one, dropped neither, and never crossed the fragility bound at the hands.

\subsubsection{Difficulty-Weighted Subtask Score}
\label{subsubsec:weighted}
The twelve subtasks are not of equal difficulty, so their success rates are not averaged. Two quantities are combined, both read from recorded bags by the same offline scorer.

The first is partial credit. Each success gate of Table~\ref{tab:fiatlux_subtasks} is declared as a list of conjuncts, so an episode that satisfies part of it can be distinguished from one that achieved nothing. Let $n$ be the number of conjuncts, $c_{\max}$ the most held simultaneously at any step, and $c_0$ the number held at reset. The gate progress is
\begin{equation}
\rho = \operatorname{clamp}\left( \frac{c_{\max} - c_0}{n - c_0}, \, 0, \, 1 \right).
\end{equation}
Conditions such as \textit{robot standing} hold at reset in every subtask, and subtracting $c_0$ keeps them from counting as progress.

The second is difficulty. Each subtask carries a weight $w$ set by the properties of the work it demands: carrying a payload, taking hold of an object at rest, releasing it so that it stays where it was put, working from the ladder, traversing the ladder, bulb and socket work, and spanning several control modes in one episode.
A subtask's score averages its success rate $\sigma$ and its mean gate progress $\bar{\rho}$, and the benchmark score is the weighted mean over the subtasks that were run:
\begin{equation}
s = \tfrac{1}{2}\sigma + \tfrac{1}{2}\bar{\rho}, \qquad S = \frac{\sum_{i} w_i s_i}{\sum_{i} w_i}
\end{equation}
An episode that never fires its gate contributes at most $0.5$ and only a fired gate reaches $1$. A subtask that was not run is excluded from both sums and reported as missing, not scored zero. Scores are not comparable across layout seeds, since the drawn layout sets the task's difficulty.

Table~\ref{tab:fiatlux_weights} gives the resulting weights. Seating the bulb from the ladder ($S_{11}$, weight $5.4$) is the heaviest subtask in the benchmark, ahead of both climbing legs that carry a payload ($S_{04}$ and $S_{10}$, $4.5$) and well ahead of the unladen ascent ($S_{02}$, $3.0$). The ordering is a design choice and records where the benchmark expects difficulty to sit.

\subsection{Domain Randomization Protocols}
Randomization runs at four moments, specified in parentheses on each term below. Layout, prestartup, and startup occur once per run, and reset recurs at every episode. The benchmark task runs all but the optional prestartup terms.

\begin{enumerate}
    \item \textbf{Scene Layout} (per layout draw): the robot, ladder, workbench with the fresh bulb, and disposal crate are each placed in their own non-overlapping floor zone, and the fixture mounts on either the ceiling or a wall. Positioning the ladder under the fixture is part of the task, so the ladder spawn zone is sampled independently of where the fixture mounts.
    \item \textbf{Light Intensity and Direction} (per reset): the lights are stage-level, so every parallel environment shares one sample. The dome light draws intensity from $\mathcal{U}(600, 1400)$ and rotates its sky texture through $[0^\circ, 360^\circ]$ about the vertical axis. The key light draws intensity from $\mathcal{U}(800, 2200)$ and is offset by $\mathcal{U}(-15^\circ, 15^\circ)$ in pitch and $\mathcal{U}(-30^\circ, 30^\circ)$ about the vertical axis.
    \item \textbf{Visual Material Tinting} (per reset): a hue--saturation--value (HSV) tint $\mathbf{T}_{\text{HSV}} \sim \mathcal{U}(0^\circ, 360^\circ) \times \mathcal{U}(0, 0.15) \times \mathcal{U}(0.7, 1.2)$ multiplies the base colors of the shared room's materials, $\mathbf{C}_{\text{render}} = \mathbf{C}_{\text{base}} \odot \mathbf{T}_{\text{HSV}}$. $\mathbf{C}_{\text{base}}$ is cached at the first tint, so resets recompute from the original and never compound.
    \item \textbf{Robot Start State} (per reset): the root pose is offset by $\pm 0.05\text{ m}$ in $x$ and $y$ and $\pm 0.1\text{ rad}$ about the vertical axis, each joint by $\pm 0.05\text{ rad}$, and all velocities begin at zero.
    \item \textbf{Grip Friction} (at startup): the hand contact material draws static friction from $\mathcal{U}(0.8, 1.2)$ and dynamic from $\mathcal{U}(0.7, 1.1)$, capped at static, so no grasp depends on the engine's $0.5/0.5$ default.
    \item \textbf{Prop Scale} (prestartup, optional): the ladder is scaled per axis, by $\mathcal{U}(0.95, 1.05)$ in $x$ and $y$ and $\mathcal{U}(0.95, 1.10)$ in $z$. The socket and the fresh bulb are scaled uniformly by $\mathcal{U}(0.9, 1.1)$ and the disposal crate is not scaled. These terms are optional to optimize for speed, as replicated environments share one geometry.
\end{enumerate}

\subsection{Automated Asset Generation}
\label{subsec:asset_gen}
To streamline how new objects enter the benchmark, Fiatlux ships a pipeline that turns images of objects into USD assets for Isaac Sim. The articulated branch runs in four stages. Hunyuan3D-2.1 \cite{hunyuan3d2025} synthesizes a textured mesh with physically based rendering (PBR) material maps from the input image. The mesh is centered at the origin and normalized to unit maximum extent. SimArt \cite{zhang2026simart} predicts a part decomposition and the kinematics connecting those parts. A USD writer then assembles the predicted articulated asset with \texttt{usd-core}. Output uses $z$ as the vertical axis and meters as units, and carries convex-decomposition collision settings, PBR textures, joint limits, and per-part density. Maximum extent and target total mass can be supplied as overrides, since SimArt's predicted size and density are approximate. Each asset must pass structural verification; benchmark use additionally requires calibration and interaction checks.

All assets used in the benchmark experiments of this paper are pre-built. The step ladder, bulb, socket, workbench, and room come from the NVIDIA Omniverse libraries, and the robot from the Unitree distribution. The pipeline is meant to widen this benchmark's object set.

Table~\ref{tab:fiatlux_assets} reports a feasibility study over six generated objects. The \emph{USD Structure} check is automated by the pipeline's verifier, which checks 27 assertions over the written stage, covering unit and up-axis conventions, mesh and texture-coordinate well-formedness, collider and physics-material binding, rigid-body structure, joint correctness, overall scale, and the external files the stage points to. The most consequential is the joint-frame coincidence check, which requires the two bodies a joint connects to place it at the same pose. All six objects pass. One trial is one fresh simulator execution, not one time step or a new object. \emph{Basic physics} asks whether the asset settles under gravity, three trials for each of the six objects. \emph{Bulb drop} releases a calibrated bulb onto a plate and requires contact; \emph{Bulb-socket interaction} drives a force-controlled insertion and extraction on each pair at a ceiling and a wall mount, using standardized contact geometry. \emph{Ladder load and push} rest 30 kg across two blocks on a settled ladder and then push laterally at 20 N, three nominal repeats and five fixed variations on each of two ladders. \emph{Scene fit} checks the geometry of two ladder, bulb and socket groups at three scales and three seeds. \emph{Articulated joint motion} drives the two predicted joints of a reused ladder.

\begin{table}[t]
\centering
\small
\caption{Feasibility validation over six generated objects.}
\label{tab:fiatlux_assets}
\begin{tabular}{l c}
\toprule
\textbf{Feasibility check} & \textbf{Passed/Tested} \\
\midrule
USD Structure & 6/6 objects \\
Basic physics & 18/18 trials \\
Bulb drop & 14/16 trials \\
Bulb-socket interaction (standardized) & 16/32 trials \\
Ladder load and push  & 16/16 trials \\
Scene fit & 12/18 configurations \\
Articulated joint motion & 0/2 joint checks \\
\bottomrule
\end{tabular}
\end{table}

\subsection{Teleoperation}
\label{subsec:teleop}
Fiatlux uses teleoperation to test whether each subtask can be completed (Section~\ref{subsec:achievability}). Each subtask provides an adapter for teleoperation, in which an operator records a take. It inherits the subtask's scene, assets, events, control rate, and success gate, so a take runs under the same physics and is scored the same way as a policy rollout.

An operator wears a PICO 4 headset with handheld controllers. Controller poses drive the arms through inverse kinematics, a trigger closes each hand, and the tracked leg motion drives the GEAR-SONIC controller \cite{luo2026sonic}. Failure terminations and the timeout are cleared so a take can continue past a mistake. The success predicate still runs, and its value is recorded for the offline scorer.

\section{Validation and Baselines}
\label{sec:results}

\subsection{Evaluation Protocol}
Evaluations use the benchmark CLI tools. Each policy runs one episode per subtask at each of four layout seeds, reported as mean and standard deviation over the seeds. Rollout trajectories are recorded as bags and scored offline, so the same recording can be re-scored under different rules without re-running the simulator.

The offline scorer reports the success rate, the clean success rate of Section~\ref{sec:method}, the mean per-episode score, and the broken and dropped rates. Peak hand contact force, mean control effort, and mean episode length are also reported. Success and the violation rates are computed independently, so a policy that seats the bulb by crushing it registers a success and a broken episode at once. The clean success rate is therefore the primary metric.

\subsection{Task Achievability}
\label{subsec:achievability}
Fiatlux tests whether each subtask can be completed in simulation by driving the robot through it under teleoperation to the success gate specified in Table~\ref{tab:fiatlux_subtasks}. Eight of the twelve subtasks have at least 10 clean full-credit teleoperation takes. For the four climbing subtasks ($S_{02}$ Climb Ladder, $S_{04}$ Descend with Bulb, $S_{10}$ Climb with Bulb and $S_{12}$ Climb Down) we test achievability by initializing the robot on the ladder steps, one at a time, holding on to the ladder, and checking for its stability.

\begin{figure*}[t]
\centering
\setlength{\tabcolsep}{1pt}
\begin{tabular}{ccccc}
\parbox[c][2.0cm][c]{0.188\textwidth}{\centering \includegraphics[height=2.0cm]{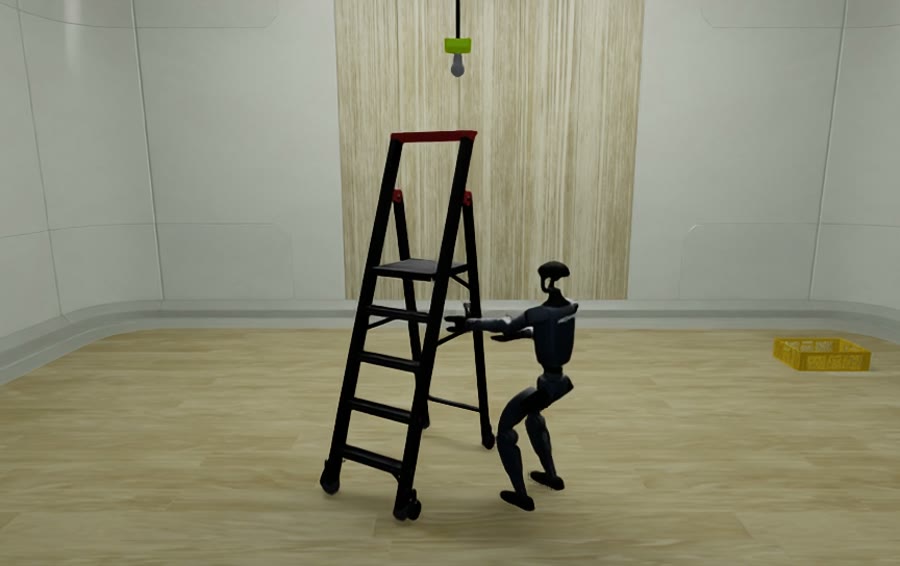}} & \parbox[c][2.0cm][c]{0.188\textwidth}{\centering \includegraphics[height=2.0cm]{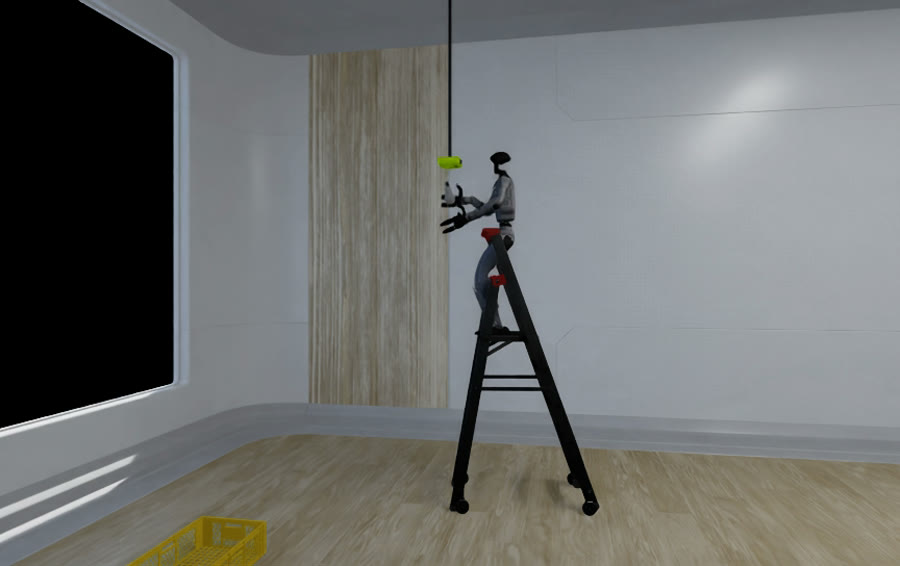}} & \parbox[c][2.0cm][c]{0.188\textwidth}{\centering \includegraphics[height=2.0cm]{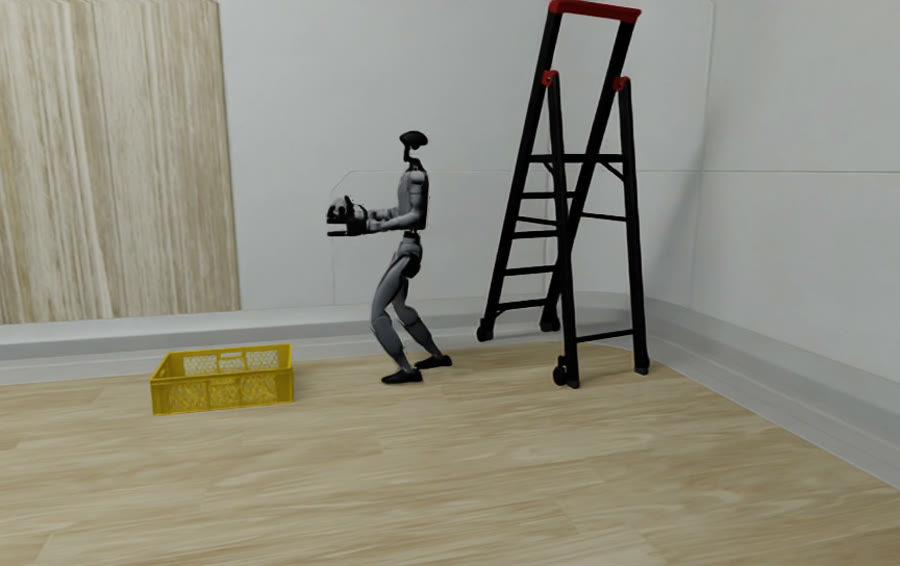}} & \parbox[c][2.0cm][c]{0.188\textwidth}{\centering \includegraphics[height=2.0cm]{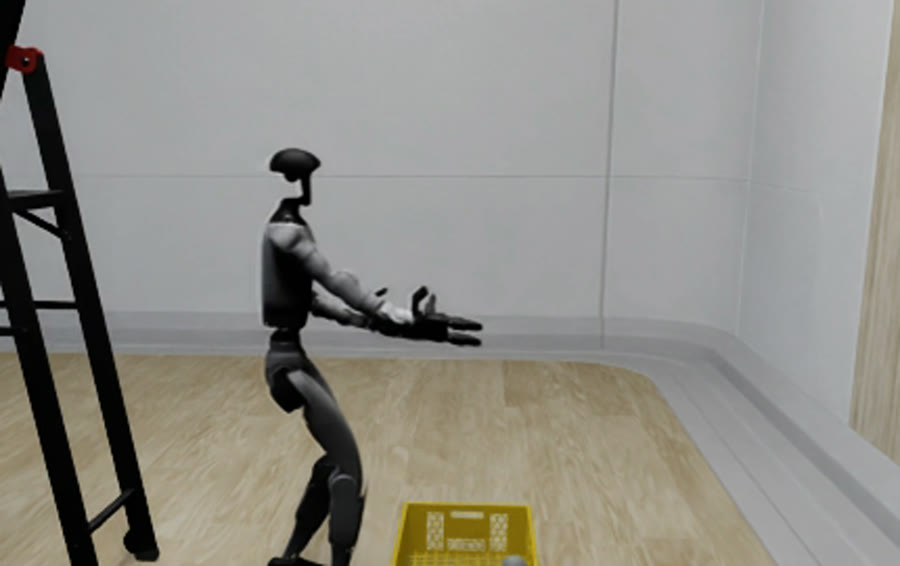}} & \parbox[c][2.0cm][c]{0.188\textwidth}{\centering \includegraphics[height=2.0cm]{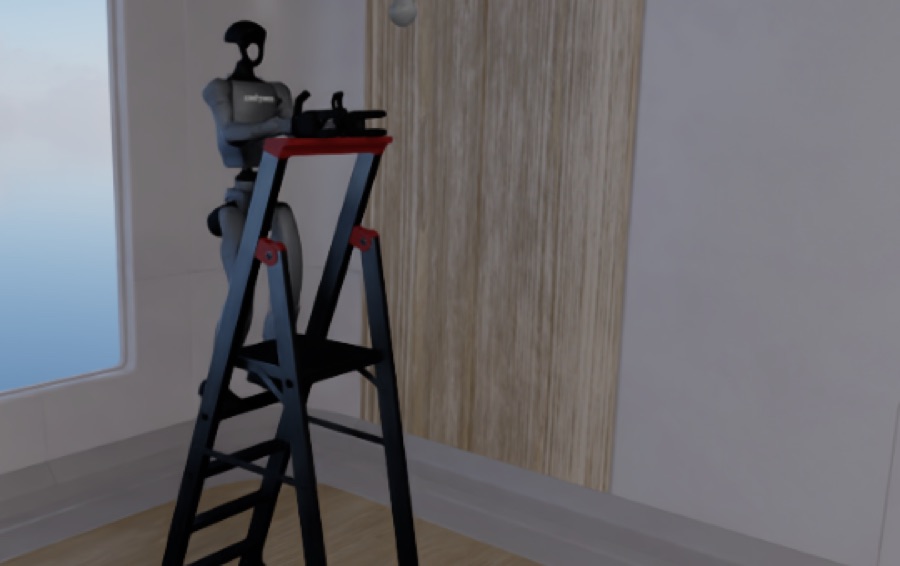}} \\
\parbox[t]{0.188\textwidth}{\centering\scriptsize $S_{01}$ Move Ladder} & \parbox[t]{0.188\textwidth}{\centering\scriptsize $S_{03}$ Remove Old Bulb} & \parbox[t]{0.188\textwidth}{\centering\scriptsize $S_{05}$ Carry Bulb to Disposal} & \parbox[t]{0.188\textwidth}{\centering\scriptsize $S_{06}$ Dispose Bulb} & \parbox[t]{0.188\textwidth}{\centering\scriptsize On-ladder stance, unladen} \\[5pt]
\parbox[c][2.0cm][c]{0.188\textwidth}{\centering \includegraphics[height=2.0cm]{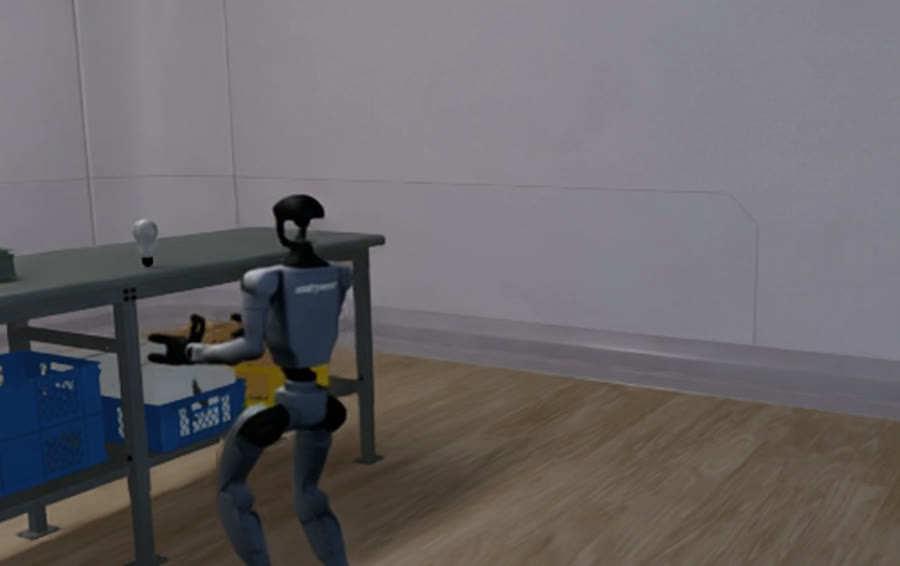}} & \parbox[c][2.0cm][c]{0.188\textwidth}{\centering \includegraphics[height=2.0cm]{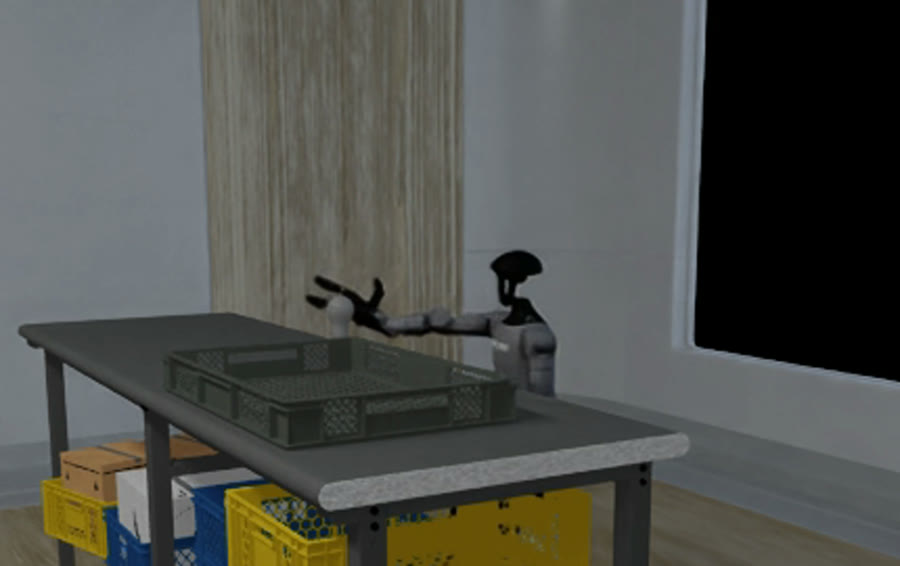}} & \parbox[c][2.0cm][c]{0.188\textwidth}{\centering \includegraphics[height=2.0cm]{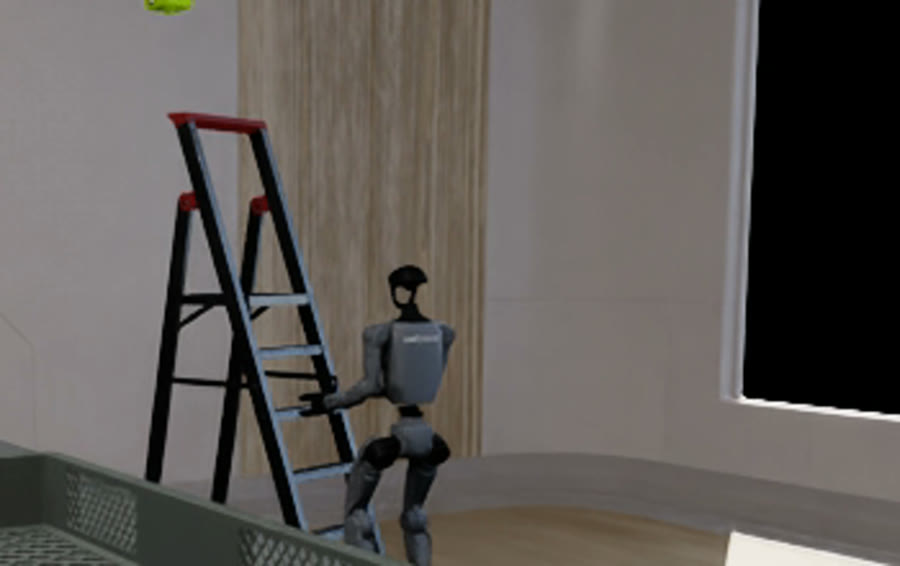}} & \parbox[c][2.0cm][c]{0.188\textwidth}{\centering \includegraphics[height=2.0cm]{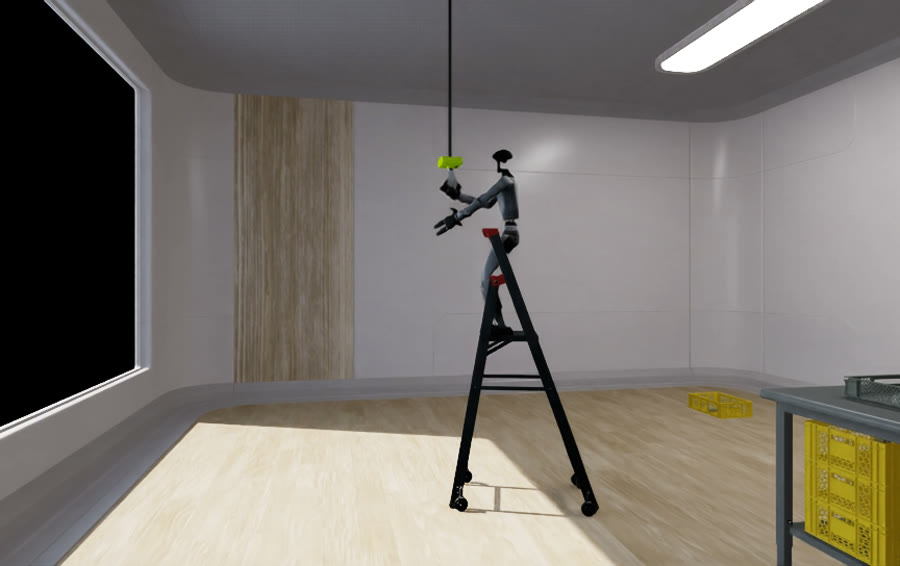}} & \parbox[c][2.0cm][c]{0.188\textwidth}{\centering \includegraphics[height=2.0cm]{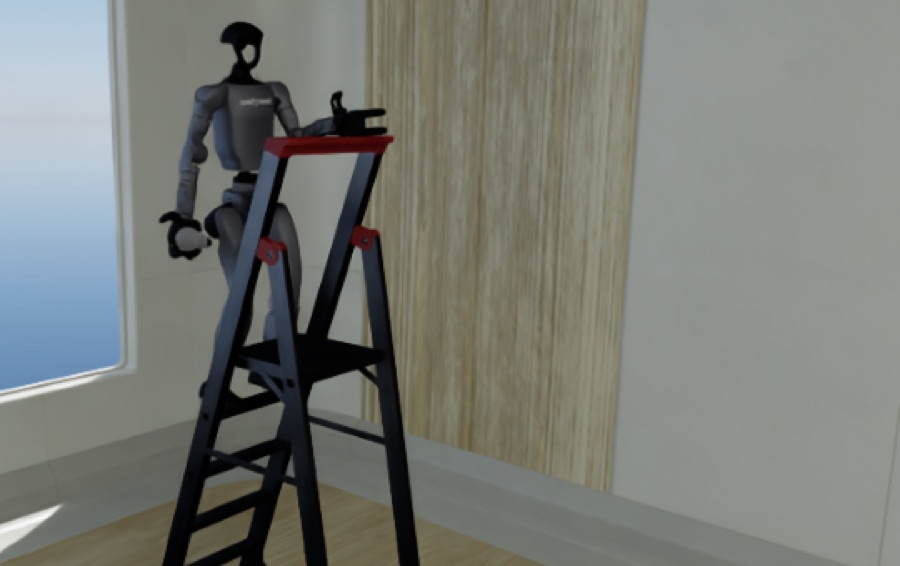}} \\
\parbox[t]{0.188\textwidth}{\centering\scriptsize $S_{07}$ Approach New Bulb} & \parbox[t]{0.188\textwidth}{\centering\scriptsize $S_{08}$ Grab New Bulb} & \parbox[t]{0.188\textwidth}{\centering\scriptsize $S_{09}$ Carry Bulb to Ladder} & \parbox[t]{0.188\textwidth}{\centering\scriptsize $S_{11}$ Screw in Bulb} & \parbox[t]{0.188\textwidth}{\centering\scriptsize On-ladder stance, laden} \\
\end{tabular}
\caption{Frames from the recorded teleoperated takes, one per subtask that has one. Each frame comes from a take that fires the subtask's gate. The four climbing subtasks are omitted: none has a take that fires its gate. The two stance frames are recorded with the robot placed on the top tread, and show that it holds the on-ladder stance with and without the bulb; the transitions to and from that stance are not demonstrated.}
\label{fig:fiatlux_takes}
\end{figure*}

A teleoperated take counts only if it satisfies every condition of the gate, sustained for the required window where one applies. Takes are recorded across multiple layout seeds, so a success cannot depend on one favorable spawn. Table~\ref{tab:fiatlux_takes} reports the takes per subtask and Fig.~\ref{fig:fiatlux_takes} shows a frame from each of the eight with a full-credit take.

\begin{table}[t]
\centering
\caption{Subtask difficulty weights and teleoperated validation takes per subtask.}
\label{tab:fiatlux_takes}
\label{tab:fiatlux_weights}
%
%
\begin{tabular}{c l c c c}
\toprule
\textbf{ID} & \textbf{Subtask} & \textbf{Weight} & \textbf{Takes} & \textbf{Full credit} \\
\midrule
$S_{01}$ & Move Ladder & 2.25 & 11 & 10 \\
$S_{02}$ & Climb Ladder & 3.00 & -- & -- \\
$S_{03}$ & Remove Old Bulb & 4.20 & 12 & 10 \\
$S_{04}$ & Descend with Bulb & 4.50 & -- & -- \\
$S_{05}$ & Carry Bulb to Disposal & 1.50 & 10 & 10 \\
$S_{06}$ & Dispose Bulb & 1.80 & 12 & 10 \\
$S_{07}$ & Approach New Bulb & 1.00 & 10 & 10 \\
$S_{08}$ & Grab New Bulb & 1.40 & 10 & 10 \\
$S_{09}$ & Carry Bulb to Ladder & 1.50 & 10 & 10 \\
$S_{10}$ & Climb with Bulb & 4.50 & -- & -- \\
$S_{11}$ & Screw in Bulb & 5.40 & 32 & 10 \\
$S_{12}$ & Climb Down & 3.00 & -- & -- \\
\bottomrule
\end{tabular}
\end{table}

\subsection{Released Baselines}
Fiatlux includes the following baselines:

\begin{enumerate}
    \item Zero Policy: Holds joint position commands at zero offset as a passive stability baseline.
    \item Random Policy: Samples each action uniformly from $[-1, 1]$.
    \item Zero-Shot GR00T N1.7: Integrates the 3B-parameter NVIDIA GR00T N1.7 model (\texttt{REAL\_G1} embodiment) with the GEAR Whole-Body Controller.
    \item Human Teleoperation: Expert demonstrations of the task execution using GEAR-SONIC Whole-Body controller and PICO 4 headset.
\end{enumerate}

\subsection{Zero-Shot VLA and Floor Baselines}
\label{subsec:vla_baseline}
The \texttt{groot} baseline, and the \texttt{zero} and \texttt{random} floor baselines against which it must be read, were each run on all twelve subtasks at four layout seeds, one episode per subtask per seed; \texttt{groot} uses the base checkpoint without fine-tuning and a per-subtask language instruction, \texttt{zero} and \texttt{random} need neither. Table~\ref{tab:fiatlux_groot} reports each policy's per-subtask score of Section~\ref{subsubsec:weighted} and the difficulty weight.

\begin{table*}[t]
\centering
\scriptsize
\caption{Per-subtask score of Zero-shot GR00T N1.7 against the \texttt{zero} and \texttt{random} floor baselines and teleoperation, twelve subtasks. Each cell is mean $\pm$ sample standard deviation over the policy's runs. GR00T/Zero/Random are averaged across 4 domain randomization seeds, Teleoperation is averaged across multiple (see Table~\ref{tab:fiatlux_takes}) recorded takes with random seeds. The climb tasks did not result in successful teleoperation. $^{*}$Teleoperation's weighted score covers only the 8 subtasks with an entry above (19.05 of 34.05 total weight); the 4 dashed subtasks are excluded from the sum, not scored zero.}
\label{tab:fiatlux_groot}
\begin{tabular}{@{}c l c c c c@{}}
\toprule
\textbf{ID} & \textbf{Subtask} & \textbf{GR00T} & \textbf{Zero} & \textbf{Random} & \textbf{Teleoperation} \\
\midrule
$S_{01}$ & Move Ladder & 0.00 & 0.00 & 0.00 & 1.00 \\
$S_{02}$ & Climb Ladder & 0.00 & 0.00 & 0.00 & -- \\
$S_{03}$ & Remove Old Bulb & 0.06$\pm$0.12 & 0.00 & 0.06$\pm$0.12 & 0.90$\pm$0.25 \\
$S_{04}$ & Descend with Bulb & 0.25 & 0.12$\pm$0.14 & 0.00 & -- \\
$S_{05}$ & Carry Bulb to Disposal & 0.25 & 0.25 & 0.00 & 1.00 \\
$S_{06}$ & Dispose Bulb & 0.00 & 0.12$\pm$0.08 & 0.33 & 0.85$\pm$0.34 \\
$S_{07}$ & Approach New Bulb & 0.00 & 0.06$\pm$0.12 & 0.06$\pm$0.12 & 1.00 \\
$S_{08}$ & Grab New Bulb & 0.00 & 0.00 & 0.00 & 1.00 \\
$S_{09}$ & Carry Bulb to Ladder & 0.25 & 0.12$\pm$0.14 & 0.00 & 1.00 \\
$S_{10}$ & Climb with Bulb & 0.19$\pm$0.12 & 0.19$\pm$0.12 & 0.00 & -- \\
$S_{11}$ & Screw in Bulb & 0.00 & 0.12$\pm$0.14 & 0.19$\pm$0.12 & 0.57$\pm$0.45 \\
$S_{12}$ & Climb Down & 0.00 & 0.00 & 0.00 & -- \\
\midrule
\multicolumn{2}{l}{\textbf{Weighted total}} & $0.0876\pm0.0261$ & $0.0861\pm0.0322$ & $0.0569\pm0.0293$ & $0.8419^{*}$ \\
\bottomrule
\end{tabular}
\end{table*}

Zero and random both stay within the same narrow, noisy band as GR00T's zero-shot result -- none of the three exceeds the others by more than seed-to-seed variation, and success rate is $0.00$ for all three on every subtask, confirmed independently from every recorded trajectory bag. Teleoperation reaches the success gate on eight of the twelve subtasks (Section~\ref{subsec:achievability}), while none of the three released baselines completes any subtask, leaving the eight achievable but unsolved and climbing not yet demonstrated.

\subsection{Failure Modes}
\label{subsec:failure_modes}
Across the three released baselines, success rate is $0.00$ on every subtask at every seed. Their non-zero entries in Table~\ref{tab:fiatlux_groot} are therefore gate progress rather than completions. The channel keeps an episode's best moment, so a failing policy banks partial credit for any gate condition it satisfies.

All three fail differently. With \texttt{random}, ten of the twelve subtasks end in a fall, none averaging more than 50 control steps. With \texttt{zero}, six do, and mean episode length runs from 38 to 6000 control steps against \texttt{random}'s 19 to 46. The robot stays upright long enough for another termination to fire first. Where the bulb is carried on or at the ladder, it is crushed or dropped before the robot falls. Elsewhere the robot neither falls nor succeeds and runs out the time budget. For teleoperation, most of the failures are connected to light bulb crushes.

Zero-shot GR00T N1.7 is not separable from \texttt{zero}, which commands no action. Their weighted scores, $0.0876 \pm 0.0261$ and $0.0861 \pm 0.0322$, differ by less than either standard deviation. No released baseline completes a subtask, so the benchmark is unsolved.

\section{Discussion and Limitations}
\label{sec:discussion}

\subsection{Whole-Body Climbing and Sim-to-Real Transferability}
Fiatlux is designed to support sim-to-real transfer onto physical Unitree G1 humanoids for locomotion and ladder climbing. The standard observation of Section~\ref{subsec:obs} carries only signals a physical robot could sense or estimate, so a policy trained on it never depends on simulator state.

The palms sense contact directly, and the feet use ankle joint torques as a proxy, since the G1 carries no foot sensor. Together these tell the policy when a limb is loaded, so it can place the next one without knowing the ladder's exact pose. The action space parameterizes joint-position targets mapped directly to low-level PD controllers at 50 Hz. A deployment adapter that bridges these targets to Unitree SDK joint commands remains future work.

\subsection{Current Benchmark Limitations}
Fiatlux has several technical limitations in its current release:

\begin{itemize}
    \item \textbf{Climbing Achievability Not Demonstrated}: The four climbing subtasks have no teleoperated take that satisfies their success gate. $S_{04}$ and $S_{12}$ stall on the descent conjunct in every recorded attempt, and no attempt at $S_{02}$ or $S_{10}$ satisfies its gate from the subtask's own start state. Clean full-credit takes exist for the eight non-climbing subtasks of twelve.
    \item \textbf{Bulb Attachment Is a State Machine, Not a Connector}: A seated bulb is held by a continuously applied wrench from a bulb attachment mechanism imitating magnetic lock, not by the socket's geometry. Modeling the connector physically remains future work.
    \item \textbf{Single-Stage RL Horizon}: Executing a complete $1{,}440\text{ s}$ horizon across navigation and ladder handling ($S_{01}, S_{05}, S_{07}, S_{09}$), climbing ($S_{02}, S_{04}, S_{10}, S_{12}$), and manipulation ($S_{03}, S_{06}, S_{08}, S_{11}$) within a single policy presents credit assignment challenges for model-free RL algorithms.
    \item \textbf{Subtask Reset Decoupling and Runtime Chaining}: The twelve subtasks are registered as independent environments, not an automated sequential chain at runtime. Each successor starts from a designed nominal spawn state, not from the terminal pose and payload the predecessor produced. Subtask policies are therefore evaluated on fixed start distributions, not on the handover distributions an upstream controller would generate.
    \item \textbf{Automated Tooling}: Automated tooling was used for code generation, for producing and checking the reported results, and for drafting and editing this paper.
    \item \textbf{Pre-Built Assets in Benchmark Environments}: The main benchmark experiments use pre-built assets from the NVIDIA Omniverse and Unitree distributions. The pipeline in Section~\ref{subsec:asset_gen} converts object images into USD assets, which are evaluated separately. Functional tests require explicit calibration, and standardized connector interfaces are evaluated as a separate assisted condition. A reused ladder’s predicted articulation failed controlled-motion checks. These results do not establish physical reconstruction accuracy, reliable articulated behavior, or fully automatic benchmark integration.
\end{itemize}

\section{Conclusion}
\label{sec:conclusion}
Fiatlux benchmarks climb-required maintenance: a Unitree G1 humanoid positions a step ladder under an overhead fixture, climbs it, exchanges a spent bulb for a fresh one, and disposes of the spent one, decomposed into twelve subtask environments scored on difficulty-weighted gates. Teleoperation establishes that eight of the twelve can be completed; the four climbing subtasks are not yet demonstrated. None of the three released baselines completes any subtask, and zero-shot GR00T N1.7 is not separable from a policy that commands no action, so every subtask the benchmark poses is open. Demonstrating the climbing subtasks and bridging joint-position targets to Unitree SDK commands are the next steps.

\section*{Acknowledgements}
This work was supported by the National Science Foundation NRT-AI 2244574 and through allocation number CIS240027 from the Advanced Cyberinfrastructure Coordination Ecosystem: Services \& Support (ACCESS) program, which is supported by National Science Foundation grants \#2138259, \#2138286, \#2138307, \#2137603, and \#2138296. The technical support and advanced computing resources from University of Hawaii Information Technology Services - Research Cyberinfrastructure, funded in part by the National Science Foundation CC* awards \#2201428 and \#2232862 are gratefully acknowledged.

This work was supported by computational resources provided by NPC Labs through B3IQ infrastructure platform.

\bibliographystyle{IEEEtran}
\bibliography{references}

\begin{thebibliography}{10}
\providecommand{\url}[1]{#1}
\csname url@samestyle\endcsname
\providecommand{\newblock}{\relax}
\providecommand{\bibinfo}[2]{#2}
\providecommand{\BIBentrySTDinterwordspacing}{\spaceskip=0pt\relax}
\providecommand{\BIBentryALTinterwordstretchfactor}{4}
\providecommand{\BIBentryALTinterwordspacing}{\spaceskip=\fontdimen2\font plus
\BIBentryALTinterwordstretchfactor\fontdimen3\font minus
  \fontdimen4\font\relax}
\providecommand{\BIBforeignlanguage}[2]{{%
\expandafter\ifx\csname l@#1\endcsname\relax
\typeout{** WARNING: IEEEtran.bst: No hyphenation pattern has been}%
\typeout{** loaded for the language `#1'. Using the pattern for}%
\typeout{** the default language instead.}%
\else
\language=\csname l@#1\endcsname
\fi
#2}}
\providecommand{\BIBdecl}{\relax}
\BIBdecl

\bibitem{makoviychuk2021isaac}
V.~Makoviychuk, L.~Wawrzyniak, Y.~Guo, M.~Lu, K.~Storey, M.~Macklin,
  D.~Hoeller, N.~Rudin, A.~Allshire, A.~Handa, and G.~State, ``Isaac gym: High
  performance gpu-based physics simulation for robot learning,'' in
  \emph{Thirty-fifth Conference on Neural Information Processing Systems
  Datasets and Benchmarks Track (Round 2)}, 2021.

\bibitem{mittal2023orbit}
M.~Mittal, C.~Yu, Q.~Yu, J.~Liu, N.~Rudin, D.~Hoeller, J.~L. Yuan, R.~Singh,
  Y.~Guo, H.~Mazhar, A.~Mandlekar, B.~Babich, G.~State, M.~Hutter, and A.~Garg,
  ``Orbit: A unified simulation framework for interactive robot learning
  environments,'' \emph{IEEE Robotics and Automation Letters}, vol.~8, no.~6,
  pp. 3740--3747, 2023.

\bibitem{yu2020meta}
T.~Yu, D.~Quillen, Z.~He, R.~Julian, K.~Hausman, C.~Finn, and S.~Levine,
  ``{Meta-World}: A benchmark and evaluation for multi-task and meta
  reinforcement learning,'' in \emph{Proceedings of the Conference on Robot
  Learning}, ser. Proceedings of Machine Learning Research, vol. 100.\hskip 1em
  plus 0.5em minus 0.4em\relax PMLR, 2020, pp. 1094--1100.

\bibitem{gu2023maniskill2}
J.~Gu, F.~Xiang, X.~Li, Z.~Ling, X.~Liu, T.~Mu, Y.~Tang, S.~Tao, X.~Wei,
  Y.~Yao, X.~Yuan, P.~Xie, Z.~Huang, R.~Chen, and H.~Su, ``Maniskill2: A
  unified benchmark for generalizable manipulation skills,''
  \emph{International Conference on Learning Representations (ICLR)}, 2023.

\bibitem{srivastava2022behavior}
C.~Li, R.~Zhang, J.~Wong, C.~Gokmen, S.~Srivastava, R.~Mart{\'i}n-Mart{\'i}n,
  C.~Wang, G.~Levine, W.~Ai, B.~Martinez, H.~Yin, M.~Lingelbach, M.~Hwang,
  A.~Hiranaka, S.~Garlanka, A.~Aydin, S.~Lee, J.~Sun, M.~Anvari, M.~Sharma,
  D.~Bansal, S.~Hunter, K.-Y. Kim, A.~Lou, C.~R. Matthews, I.~Villa-Renteria,
  J.~H. Tang, C.~Tang, F.~Xia, Y.~Li, S.~Savarese, H.~Gweon, C.~K. Liu, J.~Wu,
  and L.~Fei-Fei, ``{BEHAVIOR-1K}: A human-centered, embodied {AI} benchmark
  with 1,000 everyday activities and realistic simulation,'' \emph{arXiv
  preprint arXiv:2403.09227}, 2024.

\bibitem{nasiriany2024robocasa}
S.~Nasiriany, A.~Maddukuri, L.~Zhang, A.~Parikh, A.~Lo, A.~Joshi, A.~Mandlekar,
  and Y.~Zhu, ``{RoboCasa}: Large-scale simulation of household tasks for
  generalist robots,'' in \emph{Proceedings of Robotics: Science and Systems},
  Delft, Netherlands, July 2024.

\bibitem{gr00t2024nvidia}
{NVIDIA}, ``{GR00T N1}: An open foundation model for generalist humanoid
  robots,'' \emph{arXiv preprint arXiv:2503.14734}, 2025.

\bibitem{nasiriany2026robocasa365}
S.~Nasiriany, S.~Nasiriany, A.~Maddukuri, and Y.~Zhu, ``Robocasa365: A
  large-scale simulation framework for training and benchmarking generalist
  robots,'' in \emph{International Conference on Learning Representations
  (ICLR)}, 2026.

\bibitem{sferrazza2024humanoidbench}
C.~Sferrazza, D.-M. Huang, X.~Lin, Y.~Lee, and P.~Abbeel, ``{HumanoidBench}:
  Simulated humanoid benchmark for whole-body locomotion and manipulation,'' in
  \emph{Proceedings of Robotics: Science and Systems}, Delft, Netherlands, July
  2024.

\bibitem{gu2024humanoidgym}
X.~Gu, Y.-J. Wang, and J.~Chen, ``Humanoid-gym: Reinforcement learning for
  humanoid robot with zero-shot sim2real transfer,'' \emph{arXiv preprint
  arXiv:2404.05695}, 2024.

\bibitem{hansen2024puppeteer}
N.~Hansen, {Jyothir S V}, V.~Sobal, Y.~LeCun, X.~Wang, and H.~Su,
  ``Hierarchical world models as visual whole-body humanoid controllers,'' in
  \emph{International Conference on Learning Representations (ICLR)}, 2025.

\bibitem{hansen2024tdmpc2}
N.~Hansen, H.~Su, and X.~Wang, ``{TD-MPC2}: Scalable, robust world models for
  continuous control,'' in \emph{International Conference on Learning
  Representations (ICLR)}, 2024.

\bibitem{zhao2026ladderman}
S.~Zhao, Y.~Zhang, Z.~Lu, P.~Abbeel, R.~Duan, K.~Sreenath, Y.~Wang, C.~K. Liu,
  and G.~Shi, ``Ladderman: Learning humanoid perceptive ladder climbing,''
  \emph{arXiv preprint arXiv:2606.05873}, 2026.

\bibitem{zhao2023act}
T.~Z. Zhao, V.~Kumar, S.~Levine, and C.~Finn, ``Learning fine-grained bimanual
  manipulation with low-cost hardware,'' in \emph{Proceedings of Robotics:
  Science and Systems}, Daegu, Republic of Korea, July 2023.

\bibitem{chi2023diffusionpolicy}
C.~Chi, S.~Feng, Y.~Du, Z.~Xu, E.~Cousineau, B.~C. Burchfiel, and S.~Song,
  ``Diffusion policy: Visuomotor policy learning via action diffusion,'' in
  \emph{Proceedings of Robotics: Science and Systems}, Daegu, Republic of
  Korea, July 2023.

\bibitem{black2024pi0}
K.~Black, N.~Brown, D.~Driess, A.~Esmail, M.~R. Equi, C.~Finn, N.~Fusai,
  L.~Groom, K.~Hausman, B.~Ichter, S.~Jakubczak, T.~Jones, L.~Ke, S.~Levine,
  A.~Li-Bell, M.~Mothukuri, S.~Nair, K.~Pertsch, L.~X. Shi, L.~Smith,
  J.~Tanner, Q.~Vuong, A.~Walling, H.~Wang, and U.~Zhilinsky, ``$\pi_0$: A
  vision-language-action flow model for general robot control,'' in
  \emph{Proceedings of Robotics: Science and Systems}, Los Angeles, CA, USA,
  June 2025.

\bibitem{wang2026pointaction}
M.~Tong, H.~Jiang, Q.~Feng, L.~Liu, and J.~Gu, ``{PointAction}: {3D} points as
  universal action representations for robot control,'' \emph{arXiv preprint
  arXiv:2606.03943}, 2026, {CVPR} 2026 {4D} Vision Workshop.

\bibitem{ng1999policy}
A.~Y. Ng, D.~Harada, and S.~Russell, ``Policy invariance under reward
  transformations: Theory and application to reward shaping,'' in
  \emph{Proceedings of the Sixteenth International Conference on Machine
  Learning (ICML)}, 1999, pp. 278--287.

\bibitem{hunyuan3d2025}
{Tencent Hunyuan3D Team}, ``{Hunyuan3D 2.1}: From images to high-fidelity 3d
  assets with production-ready {PBR} material,'' \emph{arXiv preprint
  arXiv:2506.15442}, 2025.

\bibitem{zhang2026simart}
C.~Zhang, M.~Qin, Y.~Wang, B.~Xie, H.~Li, and Z.~Wang, ``{SimArt}: Decomposing
  monolithic meshes into sim-ready articulated assets via {MLLM},'' in
  \emph{SIGGRAPH Conference Papers '26}.\hskip 1em plus 0.5em minus 0.4em\relax
  Los Angeles, CA, USA: ACM, 2026, pp. 1--12.

\bibitem{luo2026sonic}
Z.~Luo, Y.~Yuan, T.~Wang, C.~Li, F.~Casta{\~n}eda, S.~Chen, Z.-A. Cao, J.~Li,
  D.~Minor, Q.~Ben, J.~Park, D.~Sami, Z.~Wang, X.~Da, R.~Ding, C.~Hogg,
  L.~Song, E.~Lim, E.~Jeong, T.~He, H.~Xue, W.~Xiao, S.~Yuen, J.~Kautz,
  Y.~Chang, U.~Iqbal, L.~J. Fan, and Y.~Zhu, ``{SONIC}: Supersizing motion
  tracking for natural humanoid whole-body control,'' \emph{Science Robotics},
  vol.~11, no. 117, p. eaed4592, 2026.

\end{thebibliography}

\end{document}